\documentclass{article}

 \usepackage[final]{nips_2016}

\usepackage[utf8]{inputenc} % allow utf-8 input
\usepackage[T1]{fontenc}    % use 8-bit T1 fonts
\usepackage{hyperref}       % hyperlinks
\usepackage{url}            % simple URL typesetting
\usepackage{booktabs}       % professional-quality tables
\usepackage{amsfonts}       % blackboard math symbols
\usepackage{amsmath}
\usepackage{amssymb}
\usepackage{amsthm}
\usepackage{nicefrac}       % compact symbols for 1/2, etc.
\usepackage{microtype}      % microtypography

\usepackage{graphicx} % more modern
\usepackage{caption}
\usepackage{subcaption}
\usepackage{algorithm}
\usepackage{algorithmic}

\usepackage{thmtools}
\usepackage{thm-restate}

\newtheorem{define}{Definition}

\newcommand{\xmark}{{\bf $\times$}}

\newboolean{proofs}
\setboolean{proofs}{false}

\title{Situational Awareness by Risk-Conscious Skills}

\author{
  Daniel J.~Mankowitz \\
  Electrical Engineering Department,\\
   The Technion - Israel Institute of Technology,\\
    Haifa 32000, Israel\\
  \texttt{danielm@tx.technion.ac.il} \\
   \And
   Aviv Tamar \\
   Electrical Engineering and \\
   Computer Sciences Department,\\
   UC Berkeley \\
   CA, USA \\
   \texttt{avivt@berkeley.edu} \\
   \AND
   Shie Mannor \\
   Electrical Engineering Department,\\
   The Technion - Israel Institute of Technology,\\
    Haifa 32000, Israel\\
   \texttt{shie@ee.technion.ac.il} \\
}

\begin{document}

\maketitle

\begin{abstract}
Hierarchical Reinforcement Learning has been previously shown to speed up the convergence rate of RL planning algorithms as well as mitigate feature-based model misspecification \cite{Mankowitz2016a,Mankowitz2016b,Bacon2015}. To do so, it utilizes hierarchical abstractions, also known as skills -- a type of temporally extended action \cite{Sutton1999} to plan at a higher level, abstracting away from the lower-level details. We incorporate risk sensitivity, also referred to as Situational Awareness (SA) , into hierarchical RL for the first time  by defining and learning risk aware skills in a Probabilistic Goal Semi-Markov Decision Process (PG-SMDP). This is achieved using our novel Situational Awareness by Risk-Conscious Skills (SARiCoS) algorithm which comes with a theoretical convergence guarantee. We show in a RoboCup soccer domain that the learned risk aware skills exhibit complex human behaviors such as `time-wasting' in a soccer game. In addition, the learned risk aware skills are able to mitigate reward-based model misspecification.
\end{abstract}

%%%%%%%%%%%%%%%%%%%%%%%%%%%%%%%%%%%%%%%%%%%%%%%%%%%%%%%%%%%%%%%%%%%%%%%%%%%%%%%
%
% The Introduction and Background
%
%%%%%%%%%%%%%%%%%%%%%%%%%%%%%%%%%%%%%%%%%%%%%%%%%%%%%%%%%%%%%%%%%%%%%%%%%%%%%%%
\section{Introduction}
\label{sec:intro}

%1. Hierarchical-RL is important for model mis-specification and other reasons.
Hierarchical-Reinforcement Learning (H-RL) is an RL paradigm that utilizes hierarchical abstractions to solve tasks. This enables an agent to abstract away from the lower-level details and focus more on solving the task at hand. Hierarchical abstractions have been utilized to naturally model many real-world problems in machine learning and, more specifically, in RL. This includes high-level controllers in robotics \cite{Peters2008,Hagras2004,daSilva2012}, strategies (such as attack and defend) in soccer \cite{Bai2015} and  video games \cite{Mann2015a}, as well as high-level sub-tasks in search and rescue missions \cite{Liu2015}. 
In RL, hierarchical abstractions are typically referred to as skills, (\cite{daSilva2012}), Temporally Extended Actions (TEAs), options (\cite{Sutton1999}) or macro-actions, (\cite{Hauskrecht1998a}). We will use the term skill to refer to hierarchical abstractions from here on in.

H-RL is important as it utilizes skills to both speed up the convergence rate in RL planning algorithms \cite{Mann2013b,Precup1997,Mann2014b} as well as mitigating model misspecification. \textbf{Model misspecification} in RL can be sub-divided into (1) \textit{feature-based} model misspecification - where a \textit{limited, sub-optimal feature set} is provided (e.g., due to limited memory resources or sub-optimal feature selection) leading to sub-optimal performance; and (2) \textit{reward-based} model misspecification whereby the reward shaping function is incorrectly designed (e.g., due to an incorrect understanding of the target problem). Previous work has focused on utilizing skills to mitigate feature-based model misspecification \cite{Mann2014b,Mankowitz2016a,Mankowitz2016b}, but have not attempted to mitigate reward-based model misspecification. \textit{Risk sensitivity} can be utilized to mitigate this form of misspecification.

%2. What is missing in Hierarchical-RL? Risk sensitivity.
%3. Why is risk sensitivity in hierarchical-RL important? Because it generates natural SA policies and mitigates reward mis-specification. This would be the first mentioning of SA.
An important factor missing in H-RL is \textit{risk sensitivity}. A risk-sensitive H-RL framework would enable us to generate skills with different Risk Attitudes, also known as Situational Awareness (SA) \cite{Endsley1995,Smith1995}, which, as we will show in our paper, allows us to mitigate \textbf{reward-based} model misspecification.  As seen in Table \ref{tbl:tea_learning_comparison}, previous work in H-RL has focused on skill learning  \cite{Mann2014b,Mankowitz2016a,Mankowitz2016b}, but has not incorporated risk-sensitivity into the H-RL objective, nor learned risk aware skills to mitigate reward-based model misspecification. From here on in, the terms risk sensitivity, risk attitude and SA will be used interchangeably.

\textbf{Situational Awareness (SA):} SA can be dependent on both time and space, although the focus of this paper is on time-based SA. We provide both definitions below.

\begin{figure}[!tbp]
  \centering
  \begin{minipage}[b]{0.45\textwidth}
   \includegraphics[width=1.0\textwidth]{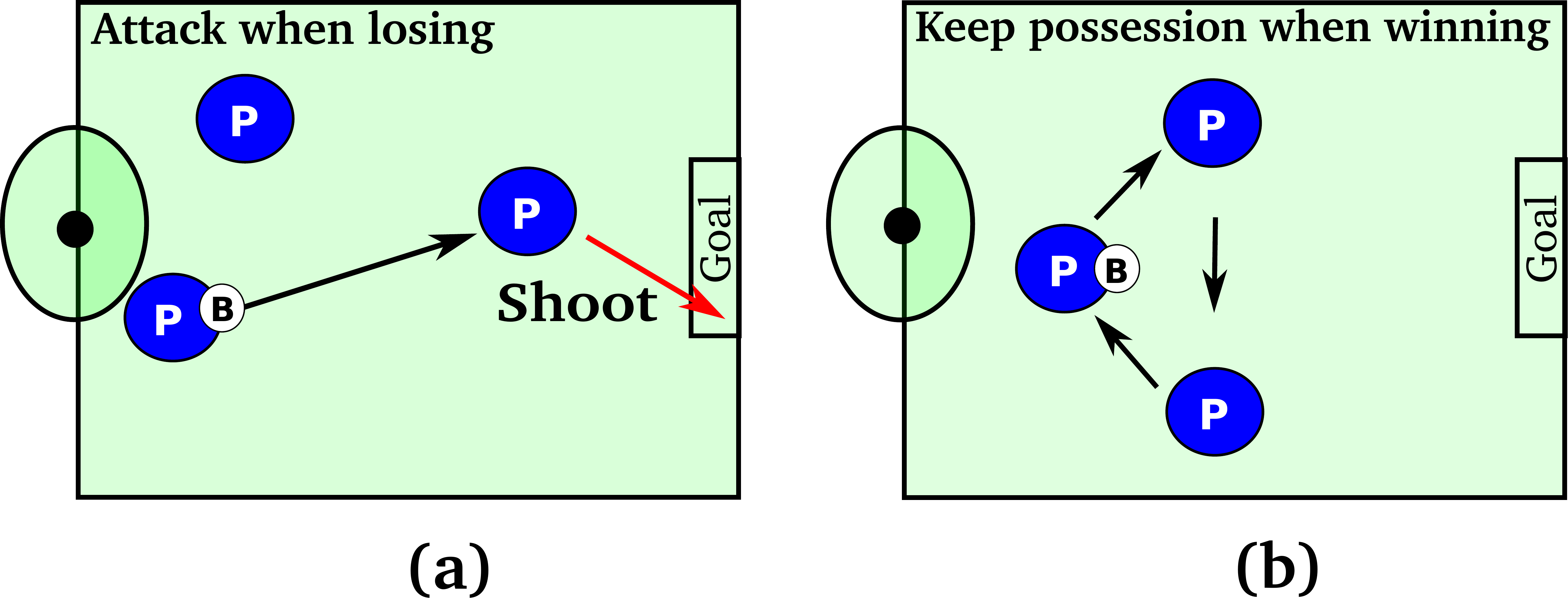}
\caption{Time-based SA - Blue players are on the same team: ($a$) Playing attacking soccer when losing a game and time is running out;  ($b$) Keeping possession and time-wasting when winning the game and time is running out.}
\label{fig:tsa}
  \end{minipage}
  \hfill
  \begin{minipage}[b]{0.45\textwidth}
    \includegraphics[width=1.0\textwidth]{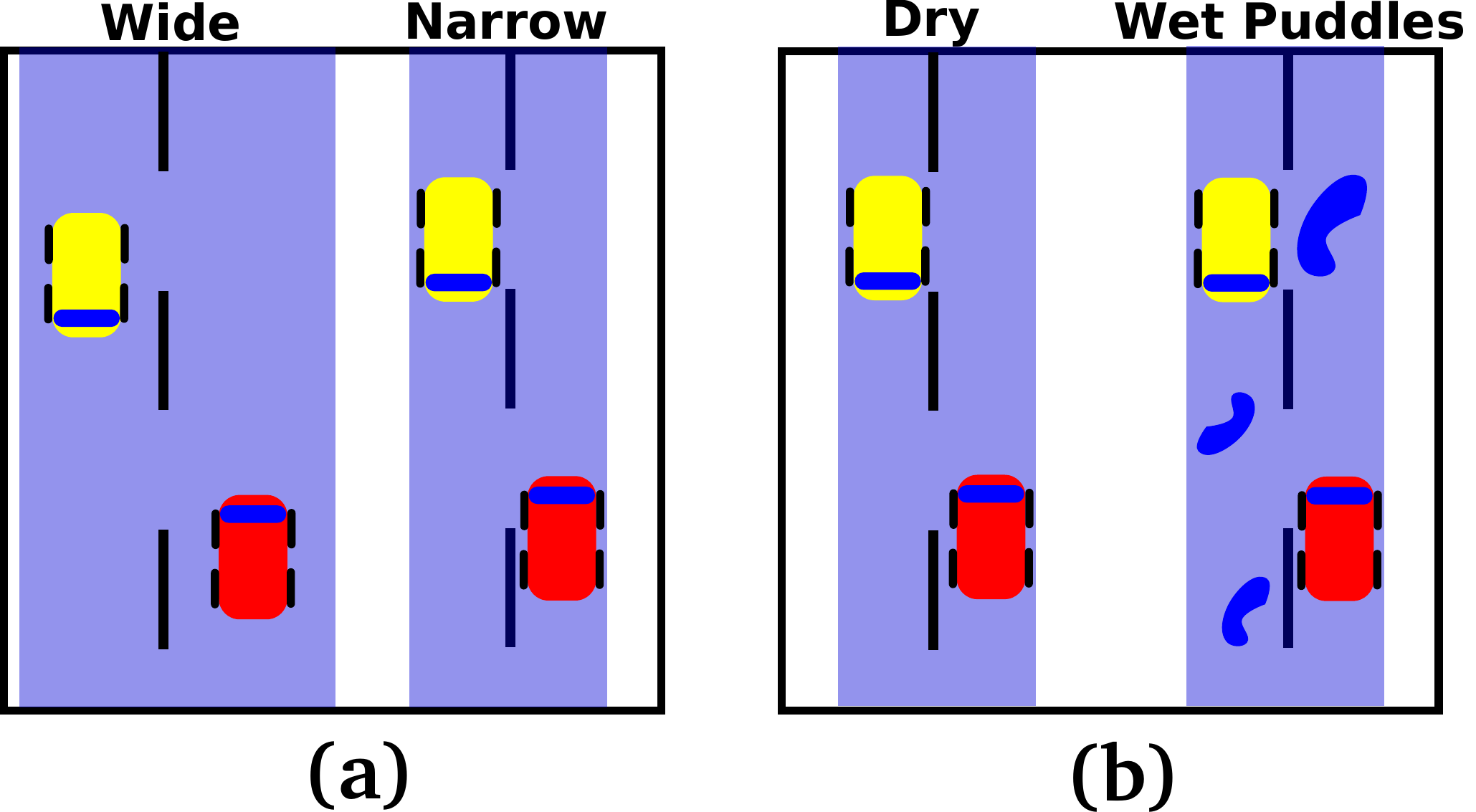}
\caption{Spatial SA: ($a$) Wide/Narrow lanes;  ($b$) Dry/Wet roads}
\label{fig:ssa}
  \end{minipage}
\end{figure}

%\begin{figure}[h!]
%\centering
%\includegraphics[width=0.4\textwidth]{timebased_sa1.pdf}
%\vspace{-0.2cm}
%\caption{Time-based SA - Blue players are on the same team: ($a$) Playing attacking soccer when losing a game and time is running out;  ($b$) Keeping possession and time-wasting when winning the game and time is running out.}
%\label{fig:tsa}
%\end{figure}

%\begin{figure}
%\centering
%\includegraphics[width=0.3\textwidth]{autonomous_car.pdf}
%\vspace{-0.2cm}
%\caption{Spatial SA: ($a$) Wide/Narrow lanes;  ($b$) Dry/Wet roads}
%\label{fig:ssa}
%\vspace{-0.5cm}
%\end{figure}

\textbf{Time-based SA:} Consider a soccer game composed of complicated strategies (skills), such as attack and defend, based on the status of the game. Consider a team losing by one goal to zero with ten minutes remaining. Here, the team needs to play \textit{attacking, risky} soccer such as making long, risky passes as well as shooting from distance to try and score goals and win the game (Figure \ref{fig:tsa}$a$). On the other hand, if the team is winning by one goal to zero with ten minutes remaining, the team needs to `waste time' by maintaining possession and playing \textit{risk-averse, defensive} football to prevent the opponent from gaining the ball and scoring goals (Figure \ref{fig:tsa}$b$). In both scenarios the team has the same objective which is to score more goals than their opponent once time runs out (I.e. win the game). Time-based SA enables an agent to act in a risk-aware manner based on the amount of time remaining in the task.

\textbf{Spatial SA:} As mentioned previously, SA can also be defined in terms of space. Consider an autonomous vehicle (the agent) driving in a narrow/wide lane or on dry/wet roads as shown in Figure \ref{fig:ssa}. The proximity of the agent to the other vehicles in the lane example (Figure \ref{fig:ssa}$a$), or the distance of the agent to other vehicles as well as puddles in the dry/wet road example (Figure \ref{fig:ssa}$b$) determines the SA and therefore the risk attitude of the agent.

%4. How do propose to add risk sensitivity to hierarchical-RL? By the PG-SMDP model and the novel SARiCOS algorithm for solving it.
Our main idea in this paper, is that a simple way to add risk-sensitivity to H-RL is by maximizing a risk-sensitive objective rather than the regular expected return formulation. One example that we focus on in this work is that of a Probabilistic Goal Markov Decision Process (PG-MDP) \cite{Xu2011}. Previous works that incorporate risk into RL have mainly been focused on learning a single risk aware policy in a non-hierarchical setting \cite{Avila1998,Tamar2015a,Tamar2015b} by maximizing the Conditional Value-at-Risk or the Value-at-Risk objectives. We provide a framework that enables an agent, for the first time to solve a task by maximizing a risk-sensitive objective in \textit{hierarchical} RL. We define a Probabilistic Goal Semi-Markov Decision Process (PG-SMDP) which naturally models this setting. By solving the PG-SMDP using our novel SARiCoS algorithm, the agent learns Risk-Aware Skills (RASs) that have a particular Risk Attitude/SA. We show that the learned risk-aware skills exhibit complex human behaviours such as \textit{time-wasting} in a soccer game. We then show in our experiments that these skills can be used to overcome reward-based model misspecification, in contrast to the regular expected return formulation. 

%5. Comparison to other work and table.
%As seen in Table \ref{tbl:tea_learning_comparison}, SARiCoS learns risk-aware skills by maximizing a risk-sensitive hierarchical RL objective and mitigates, for the first time, both forms of model misspecification. 

\begin{table*}
\scriptsize
\centering
\caption{Comparison of Approaches to SARiCoS}
\label{tbl:tea_learning_comparison}
\begin{tabular}{|c|c|c|c|c|}
\hline 
 & Maxmizes a  & Learns  & Learns  & Uses skills to\tabularnewline
 & \textbf{hierarchical risk-aware } & RL skills &  \textbf{risk-aware }RL skills & mitigate \textbf{reward-based} \tabularnewline
 & RL objective &  & (E.g. time wasting in soccer) & model misspecification\tabularnewline
\hline 
\hline 
SARiCoS (this paper) & \checkmark & \checkmark & \checkmark & \checkmark\tabularnewline
\hline 
\cite{Mankowitz2016b} & \xmark & \checkmark & \xmark & \xmark\tabularnewline
\hline 
\cite{Mankowitz2016a} & \xmark & \checkmark & \xmark & \xmark\tabularnewline
\hline 
\cite{Bacon2015} & \xmark & \checkmark & \xmark & \xmark\tabularnewline
\hline 
\cite{Masson2015} & \xmark & \xmark & \xmark & \xmark\tabularnewline
\hline 
\end{tabular}
\vspace{-0.5cm}
\end{table*}

%We evaluate our method in a simulated soccer domain, focusing on scenarios where time is running out and the agent needs to make decisions that maximize his probability of winning the game. 

\textbf{Main Contributions: } ($1$) Extending hierarchical RL to incorporate SA by defining a Probabilistic Goal Semi-Markov Decision Process (PG-SMDP) ($2$) The development of the \textbf{Si}tuational \textbf{A}wareness by \textbf{R}isk-\textbf{Co}nscious \textbf{S}kills (SARiCoS) algorithm which optimizes a hierarchical risk-aware RL objective and learns Risk-Aware Skills (RASs) that incorporate SA. ($3$) Theorem $1$ which derives a policy gradient update rule for learning Risk Aware Skills and inter-skill policy parameters in a Probabilistic Goal Semi-Markov Decision Process (PG-SMDP). ($4$) Theorem $2$ which proves that SARiCoS converges to a locally optimal solution. ($5$) Experiments in the RoboCup domain that exhibit an agent's ability to learn skills possessing SA (e.g., time wasting in a soccer game). In addition, we show the agent utilizing these skills to overcome reward-based model misspecification.

\section{Background}
\label{sec:background}
\textbf{Semi-Markov Decision Process (SMDP)} \cite{Sutton1999} A Semi-Markov Decision Process can be defined by the $5$-tuple $\langle X, \Sigma, P, R, \gamma \rangle$, where $X$ is a set of states, $\Sigma$ is a set of skills, $P$ is a transition probability function and $R$ is a bounded reward function. We assume that the rewards we receive at each timestep are bounded between $[0, R_{max}]$. Therefore $R$ forms a mapping from $X \times \Sigma$ to $[0, \frac{R_{max}}{1-\gamma}]$ and represents the expected discounted sum of rewards that are received from executing skill $\sigma \in \Sigma$ from state $x \in X$. The discount factor is defined as $\gamma \in [0,1]$. The inter-skill policy $\mu:X\rightarrow\Delta_{\Sigma}$ maps states to a probability distribution over skills. The goal in an SMDP is to find the optimal inter-skill policy $\mu^*$ that maximizes the value function $V^{\mu}(x) = \mathbb{E}\biggl[\sum_{t=0}^{\infty} \gamma^t R_t \vert x, \mu \biggr]$. This represents the expected return of following the inter-skill policy $\mu$ from state $x$. The optimal policy $\mu^*$ determines the best action to take for a given state and generates the optimal value function $V^{\mu^{*}}(s)$. 

\textbf{Skill, Option and Macro-Action} \cite{Sutton1999,daSilva2012}: An RL skill, option or macro action $\sigma$ is defined as the $3$-tuple $\sigma= \langle I, \pi_{\theta}, p(x) \rangle$ where $I$ is a set of initiation states from which a skill can be initialized or executed; $\pi_{\theta}$ is the intra-skill policy which selects the lower-level (or primitive) actions to perform whilst the skill is executing and is parameterized by $\theta \in \mathbb{R}^n$; The termination probability $p(x)$ which determines the probability of the skill terminating when in state $x$. 

\textbf{Probabilistic Goal MDP (PG-MDP)} \cite{Xu2011}: While the standard MDP objective presented above considered the expected reward, in some situations different objectives may be more appropriate. In particular, risk-sensitive criteria that maximize the probabilty of success, and not just the expected outcome, are natural objectives in domains such as finance and operations research, but also in game-playing, such as soccer. The PG-MDP is an extension of the MDP that accounts for such an objective. In a PG-MDP, the goal is to learn a policy $\pi$ that maximizes the probability that some performance threshold will be attained. That is, it aims to maximize:

\begin{equation}
\mathbb{P}(W_{\pi} \geq \beta) \enspace,
\end{equation} 
%\begin{equation}
%P(X_{\pi} \geq \beta) \geq \alpha \enspace,
%\end{equation} 

where $W_{\pi}$ is a random variable representing the total reward of the MDP under the policy $\pi$. The parameter $\beta \in \mathbb{R}$ is a performance threshold.
% and we wish to maximize the probability of getting a total reward larger than this threshold with probability $\alpha \in [0,1]$. 
The PG-MDP formulation is key for our risk shaping method, and will be further discussed when defining the PG-SMDP.

%Policy gradient
\textbf{Policy Gradient} \cite{Peters2006}: In continuous as well as high-dimensional MDPs, it is computationally inefficient to learn a policy that determines an action to perform for any given state. Policies therefore need to be \textit{generalizable}, where the policy will choose the same or similar action to perform when in nearby states. In order to achieve this generalization, a policy is parameterized using techniques such as Linear Function Approximation (LFA) (which we use in this work) \cite{Sutton1998}. A popular technique to learning the parameters for these parameterized policies is the policy gradient method. Let $J^{\pi}(\theta)$ denote the expected return of the policy parametrized by $\theta$ as $J^{\pi}(\theta) = \int_{\tau} P(\tau)R(\tau)d\tau$
where $\tau$ is a trajectory of $T$ timesteps $\langle x_1,a_1,r_1,x_2 \cdots , x_T \rangle$; $P(\tau)$ is the probability of a trajectory and $R(\tau)$ is defined as the total reward of the trajectory. Policy gradient uses sampling to estimate the gradient $\nabla_{\theta} J^{\pi}(\theta)$ and then updates the parameters using a gradient ascent update rule $\theta_{t+1} = \theta_t + \epsilon \nabla_{\theta} J^{\pi}(\theta)$ 
where $\epsilon$ denotes a positive step size.

%%%%%%%%%%%%%%%%%%%%%%%%%%%%%%%%%%%%%%%%%%%%%%%%%%%%%%%%%%%%%%%%%%%%%%%%%%%%%%%
%
% The Main Algorithm and Convergence Analysis
%
%%%%%%%%%%%%%%%%%%%%%%%%%%%%%%%%%%%%%%%%%%%%%%%%%%%%%%%%%%%%%%%%%%%%%%%%%%%%%%%
\section{Probabilistic Goal SMDP (PG-SMDP)}
\label{sec:objective}
In this work we focus on solving problems in which the agent must maximize its probability of success for solving a given task in a limited amount of time. A natural model for such problems is the the PG-MDP framework described above. However, we are interested in complex problems that require some hierarchical reasoning, and therefore propose to extend PG-MDPs to incorporate skills, leading to a PG Semi-MDP (PG-SMDP) model. We now derive an equivalent PG-SMDP with an augmented state space and skill set $\Sigma$ that can easily be utilized with policy gradient algorithms. 

We assume that we are given a set of skills $\Sigma= \{ \sigma_i \vert i=1,2, \cdots n, \sigma_j =\langle I_j, \pi_j, p_{j}(x) \rangle \}$ and \textit{inter-skill policy} $\mu(\sigma \vert x)\rightarrow \Delta_\Sigma$ which chooses a skill to execute given the current state $x\in X$. We wish to maximize the probability that the total accumulated reward, $\sum_{t=0}^T r_t$, attained during the execution of the \textit{inter-skill policy} $\mu$, passes the pre-defined performance objective threshold $\beta \in \mathbb{R}$ within $T$ timesteps.
% with probability $\alpha$. 
This takes the form of a Probabilistic Goal SMDP (PG-SMDP) (since we are incorporating skills) defined in Equation \ref{eqn:probmax}.

\begin{equation}
\max_{\mu} \mathbb{P}(\sum_{t=0}^T r_t \geq \beta | \mu) \enspace .
\label{eqn:probmax}
\end{equation}
%\begin{equation}
%\max_{\mu} P(\sum_{t=0}^T r_t \geq \beta) \geq \alpha \enspace .
%\label{eqn:probmax}
%\end{equation}

In order to solve this PG-SMDP using traditional RL techniques, we augment the state space with the total \textit{accumulated} reward \cite{Xu2011} to create an equivalent augmented PG-SMDP. We will show the important developments of this formulation for reader clarity. This will enable us to utilize traditional RL techniques in order to maximize the probability of surpassing the performance threshold $\beta$, given a set of skills $\Sigma$, within $T$ timesteps. First note that maximizing the probability can be formulated as an expectation as shown in Equation \ref{eqn:exp}.

\begin{eqnarray}\nonumber
& \max_{\mu} \mathbb{P} \left ( \left. \sum_{t=0}^T r(x_t,\sigma_t) \geq \beta \right| \mu\right) \\
= & \max_{\mu} \mathbb{E}^\mu \left [ \mathbb{I} \left( \sum_{t=0}^T r(x_t,\sigma_t) \geq \beta \right) \right]
\label{eqn:exp}
\end{eqnarray}

This expectation still contains a constraint. We now formulate an equivalent augmented PG-SMDP that removes the $\beta$ constraint and incorporates the constraint into the reward function. Define an augmented state $z = \left \{ x, w \right \}$ where $x \in X$ is the original state space and $w = \sum_{t=0}^T r(x_t,\sigma_t)$ is the accumulated reward up until time $T$. We can then define the transition probabilities in terms of the augmented state $z$ according to Equation \ref{eqn:tprobs}.

\begin{equation}
P(z' \vert z, \sigma) =\left \{ \{ x', w+r(x,\sigma)\}  \mbox{w.p } P(x' \vert x, \sigma) \right \} \enspace .
\label{eqn:tprobs}
\end{equation}

The reward function for this augmented state is then defined according to Equation \ref{eqn:rewards}. 
\begin{equation}
\tilde{r}_t(z,\sigma)= 
\begin{cases}
    0,& t < T\\
    0,& t=T, w < \beta \\
    1,& t=T, w \geq \beta
\end{cases}
\label{eqn:rewards}
\end{equation}

Together, the transition probabilities and the reward function forms an equivalent PG-SMDP with an augmented state space $z \in Z$ as shown in Equation \ref{eqn:bellmanaugmented}. This formulation learns an inter-skill policy $\mu$ that maximizes the probability that the total accumulated reward will surpass the performance threshold $\beta$ within $T$ timesteps.

\begin{equation}
\max_{\mu} \mathbb{E} \left [ \sum_{t=0}^T \tilde{r}(z_t,\sigma_t) \right]
\label{eqn:bellmanaugmented}
\end{equation}

In the next Section, we show that risk can be incorporated into the PG-SMDP by incorporating a Risk Awareness Parameter (RAP) into the typical definition of a skill to form a Risk Aware Skill (RAS). We derive a policy gradient algorithm to learn both the \textit{inter}-skill policy and the RAPs such that the agent is able to successfully solve the PG-SMDP.

\section{Risk-Aware Skill}
\label{sec:ras}
We modify the typical  definition of a skill to include a parameter, called the Risk-Awareness Parameter (RAP) $y_w \in \mathbb{R}$. This is the parameter that controls the risk-attitude of the Risk-Aware Skill (RAS). 

\begin{define}
A Risk Aware Skill (RAS) $\sigma$ is a temporally extended action that consists of the $4$-tuple $\sigma = \langle I, \pi_{\theta}, p(z), y_w \rangle$, where $I$ are the set of  states from where the RAS can be initialized; $\pi_\theta$ is the parameterized intra skill policy; $p(z)$ is the probability of terminating in state $z \in Z$;  and $y_w \in \mathbb{R}$ is the Risk-Awareness Parameter (RAP) governed by the Risk-Aware Distribution (RAD) $y_w \sim P_w(\cdot)$ with parameters $w \in \mathbb{R}^m$.
\end{define}

In practice, the RAP can parameterize the intra-skill policy, or act as a meta-parameter for the RAS (E.g. Dribble power in the RoboCup experiment (See Experiments Section)). 

%algorithm derivation
\section{SARiCoS Algorithm }
\label{sec:sars}
The Situational Awareness by Risk-Conscious Skills (SARiCoS) algorithm learns the parameters of a \textit{two-tiered} skill selection policy defined as:
\vspace{-0.1cm}
\begin{equation}
\mu_{\alpha,\Omega_{i}}(\sigma,y| z) = \mu_{\alpha}(\sigma|z)\mu_{\Omega_{i}}^{\sigma_{i}}(y|z) \enspace ,
\end{equation}

\vspace{-0.1cm}

where $\mu_{\alpha}:Z \rightarrow \Delta_\Sigma$ is the \textit{inter-skill} policy, parameterized by $\alpha \in \mathbb{R}^d$, that selects which RAS $\sigma$ needs to be executed from a set $\Sigma$ of $N$ RASs, given the current state $z \in Z$.;
$\mu_{\Omega_{i}}^{\sigma_{i}}(\cdot \vert z)$ is the RAD for RAS $\sigma_i$ with RAD parameters $\Omega_i = w_i \in \mathbb{R}^{m}$. The RAD parameters for all RASs are stored in a vector $\Omega=[\omega_1, \omega_2, \cdots, \omega_N] \in \mathbb{R}^{|N||m|\times 1}$ for algorithmic purposes.
 
The two-tiered skill selection policy is executed by first sampling a Risk-Aware Skill $\sigma_i$ to execute from $\mu_{\alpha}(\sigma|z)$. The risk attitude of the skill is then determined by sampling the RAP from the RAD $\mu_{\Omega_{i}}^{\sigma_{i}}(y|z)$. SARiCoS learns \textbf{(1)} the inter-skill policy parameters $\alpha \in \mathbb{R}^d$ and \textbf{(2)} the RAD parameters $\Omega$ to produce Situationally Aware RASs. In order to derive gradient update rules for these parameters in a policy gradient setting, we define the notion of a risk-aware trajectory.

%We now utilize a set of RAS's $\Sigma=\{\sigma_i \vert i=1,2,\cdots,n, \sigma_i=\langle I_i,\pi_i, p_i(s),y_{w_{i}} \rangle \}$, in a framework that learns \textbf{(1)} the RAD parameters $w_i$ (and implicitly the RAPs $y_{w_{i}} \sim P_{w_{i}}(\cdot)$) for each RAS $i$ and \textbf{(2)} the inter-skill parameters $\alpha \in \mathbb{R}^d$ of a parameterized inter-skill policy $\mu_{\alpha}:Z \rightarrow \Delta_\Sigma$. An \textit{inter}-skill policy selects which RAS $\sigma \in \Sigma$ needs to be executed given the current state $z \in Z$. We utilize a two-tiered selection policy that incorporates both the inter-skill policy and the RAD parameters. From here on in, RAD parameters $w_i$ for skill $i$ imply the RAPs $y_{w_{i}} \sim P_{w_{i}}(\cdot)$. In addition, the RAD parameters for each RAS are vectorized and stored in a vector $\Omega=[\omega_1, \omega_2, \cdots, \omega_N] \in \mathbb{R}^{|N||m|\times 1}$.

\textbf{Risk-Aware Trajectory:} In the standard policy gradient framework, we define a typical trajectory as $\tau = (z_t, \sigma_t, r_t, z_{t+1})_{t=0}^{T}$ where $T$ is the length of the trajectory. To incorporate the two-tiered policy into this trajectory, we define a risk-aware trajectory $\tau_r = (z_t, \sigma_t, y_{w_{\sigma_{t}}} r_t, z_{t+1})_{t=0}^{T}$ where at each timestep, we draw a RAP corresponding to the RAS $\sigma_t$ that was selected. We can therefore define the probability of a trajectory as $\mathbb{P}_{\alpha, \Omega}(\tau_r) = \mathbb{P}(z_0)\prod_{t=0}^{T-1} \mathbb{P}(z_{t+1} \vert z_t, \sigma_t) \mu_{\alpha,\Omega}(\sigma_t,y_t| z_t)$, where $P(z_0)$ is the initial state distribution; $P(z_{t+1} \vert z_t, \sigma_t)$ is the transition probability of moving from state $z_t$ to state $z_{t+1}$ given that a RAS $\sigma_t$ was executed; and $\mu_{\alpha,\Omega}(\sigma_t,y_t| z_t)$ is the two-tiered selection policy. Using this notion, it is now possible to derive the gradient update rules for each set of parameters as shown in Theorem \ref{thm:grad}.

\subsection{Inter-skill policy and RAP Update Rules} 
We define the expected reward for following a policy $\mu_{\alpha, \Omega}$:

\vspace{-0.3cm}
\begin{equation}
J(\mu_{\alpha, \Omega}) = \int_{\tau} P(\tau \vert \alpha, \Omega)R(\tau)d\tau \enspace .
\end{equation}

Let us group the parameters for the inter-skill policy and the continuous RAD Parameters into a single vector $\chi = [\alpha, \Omega] \in \mathbb{R}^{d+m\cdot N}$. 
Taking the derivative of this objective and using the well-known likelihood trick \cite{Peters2008} yields:

\vspace{-0.3cm}
\begin{equation}
\nabla_{\chi} J(\mu_{\chi}) = \int_{\tau} P(\tau \vert \chi) \nabla_\alpha \log P(\tau \vert \chi)  R(\tau)d\tau \enspace ,
\label{eqn:regPG}
\end{equation}

where $P(\tau \vert \chi)=P(z_0)\prod_{k=1}^T P(z_{k+1} \vert z_k, \sigma_k) P (\sigma_k \vert z_k, \chi)$; $z_k \in Z$ is the state at timestep $k$; $\sigma_k$ is the RAS selected at timestep $k$ and $T$ is the length of the trajectory. Since only $P (\sigma_k \vert z_k, \chi)$ is parameterized, the gradient $\nabla_\chi J(\mu_\chi)$ can be simplified to:

\vspace{-0.3cm}
\begin{equation}
\nabla_\chi J(\mu_\chi) = \int_{\tau} P(\tau \vert \chi) \nabla_\chi \log P (\sigma_k \vert z_k, \chi) R(\tau)d\tau \enspace ,
\label{eqn:generalpg}
\end{equation}

where $P (\sigma_k \vert z_k, \chi) = \mu_{\alpha}(\sigma_t|z_t)\mu_{\Omega}^{\sigma_{t}}(y_t|z_t)$. Therefore, substituting the two-tiered policy into Equation \ref{eqn:generalpg} and deriving with respect to $\alpha$ leads to the gradient update rule:

\vspace{-0.3cm}
\begin{eqnarray}\nonumber
\nabla_\alpha J(\mu_\chi) &=&  \int_{\tau} P(\tau \vert \chi) \nabla_\alpha \log \mu_{\alpha}(\sigma_t|z_t) R(\tau)d\tau \enspace .
\label{eqn:generalpg1}
\end{eqnarray}

If we represent $\mu_{\alpha}(\sigma_t|z_t)$ as a Gibb's distribution which is a common policy choice in many MDPs \cite{Sutton1998}, then we can easily derive the gradient and estimate it by samples using the following gradient update rule:

\vspace{-0.3cm}
\begin{equation}
\nabla_\alpha J(\mu_\chi) = \left <  \sum_{h=0}^H  \nabla_\alpha \log  \mu_{\alpha}(\sigma_h|z_h) \sum_{j=0}^H \gamma^j r_j     \right >
\end{equation}

If we substitute the two-tiered policy into Equation \ref{eqn:generalpg} and deriving with respect to $\Omega$ for the RAD Parameters, then we get the following gradient update rule:

\vspace{-0.3cm}
\begin{eqnarray}\nonumber
\nabla_\Omega J(\mu_\chi) &=&  \int_{\tau} P(\tau \vert \chi) \nabla_\Omega \log \mu_{\Omega}^{\sigma_{t}}(y_t|z_t) R(\tau)d\tau \enspace .
\label{eqn:generalpg2}
\end{eqnarray}

If we represent $\mu_{\Omega}^{\sigma_{t}}(y_t|z_t)$ as any distribution from the natural exponential family, then we can easily derive the gradient and estimate it by samples using the following gradient update rule:

\vspace{-0.3cm}
\begin{equation}
\nabla_\Omega J(\mu_\chi) = \left <  \sum_{h=0}^H  \nabla_\Omega \log  \mu_{\Omega}^{\sigma_{t}}(y_t|z_t) \sum_{j=0}^H \gamma^j r_j     \right >
\end{equation}

These derivations are summarized in Theorem $1$. A full proof can be found in the supplementary material.

\begin{restatable}[Gradient Update Derivation]{thm}{gradientupdate}
%\label{thm:goldbach}
%\begin{theorem}
Suppose that we are maximizing the Policy Gradient (PG) objective $J(\mu_{\alpha, \Omega}) = \int_{\tau} P_{\alpha, \Omega}(\tau)R(\tau)d\tau$ using risk-aware trajectories, generated by the two-tiered skill selection policy $\mu_{\alpha}(\sigma_t|x_t)\mu_{\Omega}^{\sigma_{t}}(y_t|z_t)$, then the expectation of the gradient update rules for the inter-skill policy parameters $\alpha \in \mathbb{R}^d$ and the RAD parameters $\Omega \in \mathbb{R}^{|N||m|}$ are the true gradients and are defined as (1) $\nabla_\alpha J(\mu_{\alpha, \Omega}) = \left <  \sum_{h=0}^H  \nabla_\alpha \log  \mu_{\alpha}(\sigma_h|z_h) \sum_{j=0}^H \gamma^j r_j     \right >$ and (2) $\nabla_\Omega J(\mu_{\alpha, \Omega}) = \left <  \sum_{h=0}^H  \nabla_\Omega \log  \mu_{\Omega}^{\sigma_{t}}(y_t|z_t) \sum_{j=0}^H \gamma^j r_j     \right >$ respectively. $H$ is the trajectory length and $<\cdot >$ is an average over trajectories as in standard PG.
\label{thm:grad}
\end{restatable}
%\end{theorem}

Given the gradient update rules, we can derive an algorithm for learning both the inter-skill parameters $\alpha \in \mathbb{R}^d$ and the continuous RAD parameters  $\Omega = [\omega_1, \omega_2, \cdots, \omega_N] \in \mathbb{R}^{|N||m| \times 1}$ for the $N$ RAS. SARiCoS learns these parameters by two timescale stochastic approximation, as shown in Algorithm \ref{alg:sars-rl}, and converges to a locally optimal solution as is proven in Theorem \ref{thm:convergence}. The convergence proof is based on standard two-timescale stochastic approximation convergence arguments  \cite{Borkar1997} and is found in the supplementary material.

\begin{restatable}[SARiCoS Convergence]{thm}{saricos}
Suppose we are optimizing the expected return $J(\mu_{\Omega,\alpha})=\intop R(\tau)P(\tau)d\tau$
for any arbitrary SARiCoS policy $\mu_{\Omega,\alpha}$ where $\Omega\in\mathbb{R}^{|N||m|}$
and $\alpha\in\mathbb{R}^{d}$ are the inter-skill and Risk Aware
Distribution parameters respectively. Then, for step sizes sequences
$\{a_{k}\}_{k=0}^{\infty},\{b_{k}\}_{k=0}^{\infty}$ that satisfy
$\sum_{k}a_{k}=\infty,\sum_{k}b_{k}=\infty,\sum_{k}a_{k}^{2}<\infty,\sum_{k}b_{k}^{2}<\infty$
and $b_{k}>a_{k}$, the SARiCoS iterates converge a.s $\alpha_k \rightarrow \alpha^{*},\Omega_k \rightarrow\bar{\lambda}(\alpha^{*})$ as $k \rightarrow \infty$
to the countable set of locally optimal points of $J(\mu_{\Omega,\alpha})$.
\label{thm:convergence}
\end{restatable} 

%The algorithm, defined as Algorithm \ref{alg:sars-rl}, learns a two-tiered policy $\mu_{\alpha, \Omega}(\sigma,y \vert z)$ by maximizing the expected return for the PG-SMDP. The Algorithm receives as input an arbitrary inter-skill policy parameterization $\alpha \in \mathbb{R}^d$ as well as a vector $\Omega \in \mathbb{R}^{|N||m|\times 1}$ of RADPs corresponding to the set of RASs. The algorithm proceeds by first fixing the RADPs and then performing policy gradient until convergence with respect to the inter-skill policy parameters. It then fixes the inter-skill policy parameters and performs policy gradient on the RADPs. This process converges to a local optimum as has been proven in the supplementary material.

\begin{algorithm}
\caption{SARiCoS Algorithm}
\label{alg:sb}
\begin{algorithmic}[1]
\REQUIRE $\alpha \in \mathcal{R}^d$. \COMMENT{Inter-skill policy parameterization}, $\Omega \in \mathcal{R}^{|N||m|\times 1}$ \COMMENT{Set of RAD parameters for each skill}
\STATE \textbf{repeat}:
\STATE $\alpha_{k+1} \rightarrow \alpha_k + a_k \nabla_{\alpha} J_{\alpha, \Omega}$ 
\STATE $\Omega_{k+1} \rightarrow \Omega_k + b_k \nabla_{\Omega} J_{\alpha, \Omega}$ \COMMENT{stepsize $b_k > a_k$}
\STATE \textbf{until convergence}
\end{algorithmic}
\label{alg:sars-rl}
\end{algorithm}

%An interesting connection can be made between parameterizing the transition probabilities and the rewards as shown in Section \ref{sec:param}, and parameterizing the skills with RSPs for a PG-MDP.  Here, fixing the RSDPs for the parameterized skill Semi-MDP yields a discrete skill Semi-MDP with parameterized transitions and rewards.

%%%%%%%%%%%%%%%%%%%%%%%%%%%%%%%%%%%%%%%%%%%%%%%%%%%%%%%%%%%%%%%%%%%%%%%%%%%%%%%
%
% Experiments and Results
%
%%%%%%%%%%%%%%%%%%%%%%%%%%%%%%%%%%%%%%%%%%%%%%%%%%%%%%%%%%%%%%%%%%%%%%%%%%%%%%%
\vspace{-0.5cm}
\section{Experiments}
\label{sec:experiments}
The experiments were performed in the RoboCup 2D soccer simulation domain \cite{Akiyama2014}; a well-known benchmark for many AI challenges. In the experiments, we demonstrate the ability of the agent to learn risk-aware skills (such as `time-wasting' in a soccer game), and therefore exhibit SA, by maximizing the PG-SMDP objective. In the RoboCup domain, we also show the agent's ability to exit local optima due to reward shaping and therefore overcome reward-based model misspecification.

%\textbf{Bottomless Pit of Death Domain:} The Bottomless Pit of Death (BPoD) domain contains two rooms, separated by a dividing wall as shown in Figure X. Below the wall is a bottomless pit of death represented by the black square. There is a wind factor in this domain that pushes the agent in a southerly direction, as shown by the arrows, with random velocity. The goal is for the agent (red ball) to circumvent the pit and reach the goal location (blue square) where it receives a large positive reward. The state space is the $x,y$ location of the agent and the action set are the cardinal directions. 

%We demonstrate the ability of the agent to be Situationally Aware. That is, to learn complex risk-aware human behaviours such as `time-wasting' in a soccer game setting. We also show examples of how SA by can overcome initially misspecified problems. 

%We also show how SA can be used to overcome initially misspecified problems.

\begin{figure*}
\centering
\includegraphics[width=1.0\textwidth]{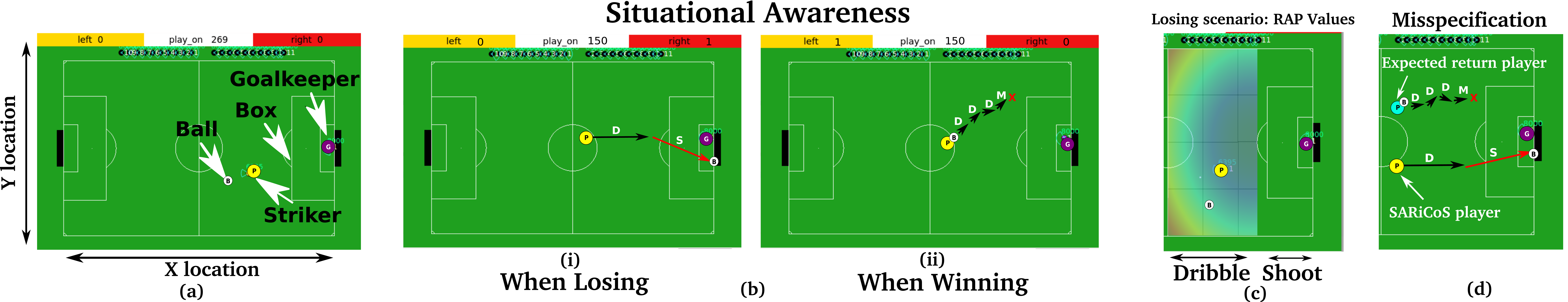}

\caption{($a$) The RO domain ($b$) SA for a winning scenario ($i$) and a losing scenario ($ii$) ($c$) The RAP values (dribble power) for the \textbf{D} skill superimposed onto the soccer field. Red indicates a fast dribble (hard kicks) and blue indicates a slow dribble (short kicks). ($d$) Reward-based model misspecification: the trajectories for the trained Expected Return (ER) and SARiCoS policies.}
\label{fig:r1}

\end{figure*}

\textbf{RoboCup Offense (RO) Domain:} This domain  \footnote{\tiny \url{https://github.com/mhauskn/HFO}} consists of two teams on a soccer field where the striker (the yellow agent) needs to score against a goalkeeper (purple circle) as shown in Figure \ref{fig:r1}$a$. The striker has $T=150$ timesteps (length of the episode) to try and score a goal. \textbf{State space} - The state space in RO consists of the continuous $\langle x,y \rangle$ field locations of the striker, ball, goalposts and goalkeeper as well as the cumulative sum of rewards $w$. \textbf{Skills} - The Risk-Aware Skill (RAS) set $\Sigma$ in each of the experiments consists of three RAS: (1) Move to the ball \textbf{(M)}, (2) Move to the ball and shoot towards the goal \textbf{(S)} and (3) Move to the ball and dribble in the direction of the goal \textbf{(D)}. Each RAS $i$ is parameterized with a Risk Aware Parameter $y_{w_{i}}$.  We focus on learning the dribbling power RAP $y_{w, D}$ that controls how hard the agent kicks the ball when performing the skill Dribble. \textbf{Data:} SARiCoS is trained over $3$ independent trials with $20,000$ episodes per trial. \textbf{Learning Algorithm and features} - The learning algorithm for both the inter-skill policy parameters $\alpha$ and the RAPs for the RASs is Actor Critic Policy Gradient (AC-PG) \footnote{AC-PG has lower variance compared to regular PG and the convergence guarantees are trivial extensions of the current proof.}. The inter-skill policy $\mu$ that chooses which RAS to execute is represented by a Gibb's distribution with Fourier Features. The Risk Aware Distribution (RAD) is represented as a normal distribution $y \sim \mathcal{N}(\phi(s)^T \omega, V)$ with a fixed variance $V$. Here,  $\phi(s)$ are state dependent features $[1,x_{agent},y_{agent},w,distGoal]$ representing the agent's $x,y$ location, the cumulative reward and the distance of the agent to the goal. \textbf{Rewards} - Engineering of the reward in RL is common practice for the RoboCup domain \cite{Hausknecht2015,Bai2015}. The rewards for both of the RoboCup scenarios have been engineered based on logical soccer strategies. The striker gets small positive rewards for dribbling outside the box $r_{D,far}$ and shooting when inside or near the box $r_{S,near}$. Negative rewards come about when the striker dribbles inside the box, $r_{D,near}$, or shoots from far, $r_{S,far}$, as the striker has a smaller probability of scoring \cite{Yiannakos2006}. The striker also gets a small positive reward for moving towards the ball $r_{move}$. There is also a game score reward $r_{score}$, which is positive if winning and negative if losing or drawing. In the PG-SMDP setting, the rewards $\tilde{r}=1$ if $w>=\beta$ at the end of each episode, otherwise the reward is $0$ at each timestep. In the Expected Return setting (see Reward-Based Model Misspecification), the regular rewards are utilized at each timestep.

%\begin{figure*}
%\centering
%\includegraphics[width=1.0\textwidth]{../figures/robocup1_overlay_episode_workshop_nips.pdf}
%\caption{($a$) The RO domain ($b$) SA in a Losing scenario: The RAP as a function of the agent's $x$ location on the field. ($c$) SA for a winning scenario ($i$) and a losing scenario ($ii$) ($d$) A superposition of the RAP values for the Dribble skill on the soccer field. Red indicates a fast dribble (hard kicks) and blue indicates a slow dribble (short kicks). ($e$) The average number of steps taken to complete an episode for the winning and losing scenarios}
%\label{fig:r1}
%\vspace{-0.6cm}
%\end{figure*}

\subsection{Situational Awareness by Risk-Conscious Skills}
\label{sec:sa}
In this section we show that learning the inter-skill policy and RAD parameters using SARiCoS so as to maximize a PG-SMDP can bring about risk-aware skills that exhibit time-based SA. We provide the agent with two different soccer situations: \textbf{(1)} The agent is losing the game $0-1$;  \textbf{(2)} The agent is winning the game $1-0$. Similar results are obtained for different scores (e.g., $2-0$ and $0-2$ etc. and have therefore been omitted). For all of the scenarios, the performance threshold $\beta$ for the PG-SMDP is set to a constant value ($\beta=1.0$) \textit{a-priori}.

%This is performed in the RoboK domain. 

%\begin{figure}
%\centering
%\includegraphics[width=0.5\textwidth]{../figures/misspecified/scenarios.pdf}
%\caption{Situational Awareness strategies learned by an agent when ($a$) winning, ($b$) drawing and ($c$) losing a soccer game. ($d$) A sub-optimal solution when winning resulting in model misspecification. See the \textbf{video} in the supplementary material.}
%\label{fig:scenario}
%\end{figure}

\textbf{\underline{SA in a Losing Scenario:}} In a scenario where a team is losing and time is running out, the team needs to play risky, attacking soccer to try and score goals. The agent is placed in a losing scenario where the score  is $0-1$ to the opposition with $150$ timesteps remaining. Using SARiCoS, the agent learns to perform a fast Dribble by kicking the ball with significant power to make quick progress along the pitch and get in a position to shoot for goal as seen in Figure \ref{fig:r1}$b(i)$. The average RAP value for the Dribble RAS is approximately $100$ (max value $150$, min value $0$) prompting the agent to kick the ball with significant power and quickly advance up the pitch. The RAP is state dependent, enabling the agent to learn to initially kick the ball with a large amount of power when near the half-way line and then decrease the dribble power when approaching the goal so as to prevent losing possession to the goalkeeper. This is seen from the dribble power color gradient superimposed onto the RO domain in Figure \ref{fig:r1}$c$. The color gradient varies from powerful kicks (in red)  to soft kicks (in blue). Once the agent is near the goal, it executes the skill \textbf{Shoot} as seen in the figure \ref{fig:r1}$b(i)$. The average episode length is $70.0\pm1.0$ (mean$\pm$std) as seen in Table \ref{tab:capt} and the average number of goals scored over $100$ evaluation episodes is $74.3\pm6.5$. In addition, the keeper captures the ball on average  $21\pm 5.29$ times indicating that the striker is playing risky football with aggressive dribbling and, as a result, scores a high number of goals. In addition, the average reward is consistently higher than the $\beta$ threshold.

%\textbf{\underline{SA in a Drawing Scenario:}} In a drawing scenario, a team wants to win the game and therefore plays attacking soccer, but does not take as many risks as in a losing scenario, since a team still gets a point for a draw in a regular soccer league. Using SARiCoS, the agent learns this slightly more risk-averse behavior as shown in \ref{fig:scenario}$b$. Figure \ref{fig:r1}$e$  reiterates this point as the striker takes more time ($119$ steps on average) than in the losing scenario to score goals, as expected. 

\textbf{\underline{SA in a Winning Scenario:}} When winning a game with little time remaining, a natural strategy is to hold onto the ball and \textit{run out the clock \footnote{\tiny \url{http://www.collinsdictionary.com/dictionary/american/run-out-the-clock}}} (\textbf{`time-wasting'}) so as to prevent the opposing team from gaining possession and possibly scoring a goal. SARiCoS learns `time-wasting' since when the agent is winning the game $1-0$, the agent slowly dribbles his way up the pitch, collecting the dribble from far rewards $r_{D,far}$ in the process as seen in Figure \ref{fig:r1}$b(ii)$. Once the agent crosses the performance threshold, it stands on the ball,  and \textit{wastes} time by executing the \textbf{M} skill, whilst continuing to collect the positive score rewards $r_{score}$ and the small positive $r_{move}$ rewards. This strategy causes the agent to take the largest amount of time on average ($142.3\pm1.5$ steps) to complete each episode (time wasting) and as a result only scores $1.3\pm0.6$ goals. However, the ball is almost never captured by the opponent $7.3\pm0.6$ times on average per $100$ evaluation episodes.  See a \textbf{Video}\footnote{\tiny\url{https://youtu.be/xA-8rWJ4a7I}} of the agent's behavior in each of these scenarios.

\vspace{-0.2cm}
\subsection{Mitigating Reward-based Model Misspecified}
\label{sec:misspecified}
%The previous experiments have demonstrated the ability to learn risk-aware skills that exhibit complex, SA behaviors such as 'time-wasting'.

The learned risk-aware skills can be utilized to overcome reward-based model misspecification. We focus on the losing scenario in RoboCup soccer. We compared SARiCoS to the regular Expected Return (ER) formulation, i.e., an implementation of Actor-Critic Policy Gradient that utilizes regular rewards at each timestep to learn a game-winning policy. 

As seen in Figure \ref{fig:r1}$d$, the ER striker (light blue circle) does not learn to score goals as the algorithm settles quickly on collecting positive dribble from far rewards $r_{D,far}$ and moving to the ball rewards $r_{move}$. The ER agent therefore gets stuck in a local optima causing the agent to execute \textbf{D} until it settles on the \textbf{M} skill and stands on the ball, receiving small positive rewards. As seen in Table \ref{tab:capt}, the ER agent only manages to score $1.7\pm1.2$ goals on average and has a low average reward $-0.3\pm0.1$, well below the $\beta$ threshold.

These rewards are therefore not enough to enable the SARiCoS agent (yellow circle in Figure \ref{fig:r1}$d$) to pass its performance threshold $\beta$, especially since, in the losing scenario, the agents are also receiving a negative game score reward $r_{score}$ at each timestep. This forces the SARiCoS agent to search for additional rewards such as a goal-scoring reward. As seen in Table \ref{tab:capt}, the SARiCoS agent learns to score goals ($74\pm6.5$), and achieves average reward well above the $\beta$ performance threshold. As a result it mitigates the reward shaping-based model misspecification.

\begin{table}
\centering
\caption{Performance of the trained SARiCoS and Expected Return (ER) policies in the winning/losing scenarios averaged over $100$ evaluation episodes.}
\small
\begin{tabular}{|c|c|c|c|}
\hline 
 & \textbf{SARiCoS} & \textbf{SARiCoS} & \textbf{ER}\tabularnewline
 & Winning & Losing & Losing\tabularnewline
\hline 
\hline 
Goals & $1.3\pm0.6$ & \textbf{74.3$\pm$6.5} & \textbf{1.7$\pm$1.2}\tabularnewline
\hline 
Out of Time & \textbf{90.3$\pm$1.5} & $1.0\pm1.7$ & $47.0\pm3.6$\tabularnewline
\hline 
Avg Reward & $3.9\pm1.1$ & \textbf{6.3$\pm$0.2} & \textbf{-0.3$\pm$0.1}\tabularnewline
\hline 
Episode Length & $142.3\pm1.5$ & $70.0\pm1.0$ & $107.3\pm3.8$\tabularnewline
\hline 
\end{tabular}
\label{tab:capt}
\end{table}

%\begin{figure}
%\centering
%\includegraphics[width=0.2\textwidth]{../figures/reward-misspecification/misspecification.pdf}
%\caption{Losing Scenario: \textbf{To do} The learning curve of the SARiCoS policy compared to the regular expected return formulation.}
%\label{fig:misspecified}
%\vspace{-0.2cm}
%\end{figure}

%($b$) RoboCup2: The standard expected return for the winning scenario vs the PG-SMDP return

\vspace{-0.3cm}
\section{Discussion}
%Remember to mention 
We have defined a PG-SMDP which provides a natural risk-sensitive objective for learning SA in hierarchical RL. We find it interesting that an agent can learn a complex human behavior by simply maximizing a risk-sensitive objective. To do so, we have introduced Risk-Aware Skills (RASs) --- a type of parameterized option \cite{Sutton1999} with an additional Risk-Aware Parameter (RAP). We have developed the Situational Awareness by Risk-Conscious Skills (SARiCoS) algorithm which learns both the inter-skill policy that chooses RASs to execute, as well as learning the RAPs for each RAS. We have shown that this algorithm converges to a locally optimal solution. We also show that SARiCoS can induce situational awareness (E.g. `time-wasting') in Risk-Aware Skills in a time dependent RoboCup soccer scenario.  In principle, any other risk criteria can be incorporated into this work such as exponential risk, CVaR and VaR \cite{Avila1998,Tamar2015a,Tamar2015b}. Extensions of this work include optimizing a PG-MDP performance threshold $\beta$ for each RAS as well as utilizing SA in lifelong learning problems \cite{Thrun1995,Pickett2002,Brunskill2014}. The SARiCoS policy could also be implemented as a Deep Network \cite{Mnih2015}, leading to more complex policies on higher dimensional problems.

\section*{Acknowledgements}
\label{sec:acknowledgement}
The research leading to these results has received funding from the European Research Council under the European Union's Seventh Framework Program (FP/2007-2013) / ERC Grant Agreement n. 306638.

\bibliography{tmann}
\bibliographystyle{icml2016}

\newpage
\appendix

\section{SARiCoS Supplementary Material}

\subsection{Full Derivation of Theorem 1}
We define the expected reward for following a policy $\mu_{\alpha,\Omega}$ as:

\begin{equation}
J(\mu_{\alpha, \Omega}) = \int_{\tau} P(\tau \vert \alpha, \Omega)R(\tau)d\tau \enspace .
\end{equation}

Let us group the parameters for the inter-RAS policy and the continuous RADPs into a single vector $\chi = [\alpha, \Omega] \in \mathbb{R}^{d+m\cdot N}$. 
Taking the derivative of this objective and using the well-known likelihood trick \cite{Peters2008} yields:

\begin{eqnarray}
\nabla_{\chi} J(\mu_{\chi}) &=& \nabla_{\chi} \int_{\tau} P(\tau \vert \chi)   R(\tau)d\tau\\
&=&  \int_{\tau}\nabla_{\chi} P(\tau \vert \chi)   R(\tau)d\tau\\
&=& \int_{\tau} P(\tau \vert \chi) \nabla_\alpha \log P(\tau \vert \chi)  R(\tau)d\tau \enspace ,
\label{eqn:regPG}
\end{eqnarray}

where $P(\tau \vert \chi)=P(z_0)\prod_{k=1}^T P(z_{k+1} \vert z_k, \sigma_k) P (\sigma_k \vert z_k, \chi)$ where $z_k \in Z$ is the state at timestep $k$; $\sigma_k$ is the RAS selected at timestep $k$ and $T$ is the length of the trajectory. Since only $P (\sigma_k \vert z_k, \chi)$ is parameterized, the gradient $\nabla_\chi J(\mu_\chi)$ can be simplified as follows:

\begin{eqnarray}
\nabla_{\chi} J(\mu_{\chi}) &=& \int_{\tau} P(\tau \vert \chi) \nabla_\chi \log P(\tau \vert \chi)  R(\tau)d\tau\\
&=& \int_{\tau} P(\tau \vert \chi) \nabla_\chi \log \biggl[P(z_0)\prod_{k=1}^T P(z_{k+1} \vert z_k, \sigma_k) P (\sigma_k \vert z_k, \chi) \biggr]  R(\tau)d\tau\\
&=& \int_{\tau} P(\tau \vert \chi) \nabla_\chi \biggl[ \log P(z_0)+\Sigma_{k=1}^T \log P(z_{k+1} \vert z_k, \sigma_k)+ \log P (\sigma_k \vert z_k, \chi)  \biggr] R(\tau)d\tau\\
&=& \int_{\tau} P(\tau \vert \chi) \nabla_\chi  \log P (\sigma_k \vert z_k, \chi)  R(\tau)d\tau
\label{eqn:generalpg}
\end{eqnarray}

where $P (\sigma_k \vert z_k, \chi) = \mu_{\alpha}(\sigma_t|z_t)\mu_{\Omega}^{\sigma_{t}}(y_t|z_t)$. Therefore, substituting the two-tiered policy into Equation \ref{eqn:generalpg} and deriving with respect to $\alpha$ leads to the gradient update rule:

\begin{eqnarray}\nonumber
\nabla_\alpha J(\mu_\chi) &=&  \int_{\tau} P(\tau \vert \chi) \nabla_\alpha \log \mu_{\alpha}(\sigma_t|z_t) R(\tau)d\tau \enspace .
\label{eqn:generalpg1}
\end{eqnarray}

If we represent $\mu_{\alpha}(\sigma_t|z_t)$ as a Gibb's distribution which is a common policy choice in many MDPs \cite{Sutton1998}, then we can easily estimate the gradient by sampling:

\begin{equation}
\nabla_\alpha J(\mu_\chi) = \left <  \sum_{h=0}^H  \nabla_\alpha \log  \mu_{\alpha}(\sigma_h|z_h) \sum_{j=0}^H \gamma^j r_j     \right > \enspace ,
\end{equation}

where $H$ is the length of a trajectory; and $\biggl<\cdot \biggr>$ represents an average over trajectories. If we derive Equation \ref{eqn:generalpg} with respect to $\Omega$ for the RADPs, then we get the following gradient update rule:

\begin{eqnarray}\nonumber
\nabla_\Omega J(\mu_\chi) &=&  \int_{\tau} P(\tau \vert \chi) \nabla_\Omega \log \mu_{\Omega}^{\sigma_{t}}(y_t|z_t) R(\tau)d\tau \enspace .
\label{eqn:generalpg2}
\end{eqnarray}

If we represent $\mu_{\Omega}^{\sigma_{t}}(y_t|z_t)$ as any distribution from the natural exponential family, then we can easily estimate the gradient by samples using the following gradient update rule:

\begin{equation}
\nabla_\Omega J(\mu_\chi) = \left <  \sum_{h=0}^H  \nabla_\Omega \log  \mu_{\Omega}^{\sigma_{t}}(y_t|z_t) \sum_{j=0}^H \gamma^j r_j     \right > \enspace .
\end{equation}

These derivations are summarized in Theorem $1$.

\gradientupdate*

\subsection{Proof of Theorem 2: SARiCoS Convergence}

\saricos*

The true gradient of the two-tiered policy $\mu_{\Omega,\alpha}$
is:

\[
\nabla_{\Omega,\alpha}J(\mu_{\Omega,\alpha})=\mathbb{E}[R_{\tau}\nabla\log P(\tau)]
\]

where $R_{\tau}=\sum_{t=0}^{h-1}\gamma^{t}r_{t}$ is the discounted
cumulative reward for a trajectory $\tau$ of length $h$; the term
$P(\tau)=P(x_{0})\Pi_{i=0}^{h-1}P(x_{i+1}\vert x_{i},\sigma_{i})\mu_{\Omega,\alpha}(\sigma_{i},y_{i}\vert x_{i})$
is the probability of a trajectory for a given policy $\mu_{\Omega,\alpha}(\sigma_{i},y_{i}\vert x_{i})$. 

The estimated gradient is:

\begin{eqnarray*}
\hat{\nabla}J(x) & = & R_{\tau}\nabla\log\mu_{\Omega,\alpha}(\sigma_{t},y_{t}\vert x_{t})\\
 & = & R_{\tau}\nabla\log(\mu_{\alpha}(\sigma_{t}\vert x_{t})\mu_{\Omega}^{\sigma_{t}}(y_{t}\vert x_{t}))
\end{eqnarray*}

We need to prove that the parameters $\alpha\in\mathbb{R}^{d}$ of
the inter-skill policy and the risk-aware parameters $\Omega\in\mathbb{R}^{|N||m|}$
converge to a locally optimal solution. Here, $N$ is the number of
skills and $m$ is the number of risk-aware distribution parameters
for each skill. In order to do so, we first derive the gradient with
respect to $\alpha$ to yield the following recursive update equations:

\begin{eqnarray}
\alpha_{k+1} & = & \Gamma_{\alpha}(\alpha_{k}+a_{k}\hat{\nabla}_{\alpha}J(x))\nonumber \\
 & = & \Gamma_{\alpha}(\alpha_{k}+a_{k}(R_{\tau}\nabla_{\alpha}\log\mu_{\alpha}(\sigma_{t}\vert x_{t})))\nonumber \\
 & \overset{(1)}{=} & \Gamma_{\alpha}(\alpha_{k}+a_{k}(R_{\tau}z_{k}^{\alpha}))\nonumber \\
 & = & \Gamma_{\alpha}(\alpha_{k}+a_{k}(R_{\tau}z_{k}^{\alpha}-\mathbb{E}[R_{\tau}z_{k}^{\alpha}]+\mathbb{E}[R_{\tau}z_{k}^{\alpha}]))\nonumber \\
 & = & \Gamma_{\alpha}(\alpha_{k}+a_{k}(f(\alpha(k),\Omega)+N_{k+1}))\label{eq:fast}
\end{eqnarray}
\\
where $(1)$ $z_{k}^{\alpha}=\nabla_{\alpha}\log\mu_{\alpha}(\sigma_{t}\vert x_{t})$
and $N_{k+1}=R_{\tau}z_{k}^{\alpha}-\mathbb{E}[R_{\tau}z_{k}^{\alpha}]$
is a zero-mean martingale difference sequence; $f(\alpha(k),\Omega)=\mathbb{E}[R_{\tau}z_{k}^{\alpha}]$
and $\Gamma_{\alpha}:\mathbb{R}^{d}\rightarrow\mathbb{R}^{d}$ is
a projection operator that projects any $\alpha_{k}$ to a compact
region $C=\{\alpha\vert g_{i}(\alpha)\leq0,i=1,\cdots l\}\in\mathbb{R}^{n}$
where $g_{i}(\cdot),i=1,\cdots l$ represent the continuously differentiable
constraints that project the iterates to a compact region defined
by a ball with a smooth boundary. This operator ensures that the iterates
remain bounded. It can be seen by inspection that this recursion represents
a noisy discretization of the Ordinary Differential Equation (ODE)
\cite{Borkar1997}:

\[
\dot{\alpha}=\Gamma_{\alpha}(\mathbb{E}[R_{\tau}z_{k}^{\alpha}])
\]

We also derive the recursive update for the risk-aware parameters
$\Omega$ as follows:

\begin{eqnarray}
\Omega_{k+1} & = & \Gamma_{\Omega}(\Omega_{k}+b_{k}\hat{\nabla}_{\Omega}J(x))\nonumber \\
 & = & \Gamma_{\Omega}(\Omega_{k}+b_{k}(R_{\tau}\nabla\log\mu_{\Omega}^{\sigma_{t}}(y_{t}\vert x_{t}))\nonumber \\
 & = & \Gamma_{\Omega}(\Omega_{k}+b_{k}(R_{\tau}z_{k}^{\Omega}))\nonumber \\
 & = & \Gamma_{\Omega}(\Omega_{k}+b_{k}(R_{\tau}z_{k}^{\Omega}-\mathbb{E}[R_{\tau}z_{k}^{\Omega}]+\mathbb{E}[R_{\tau}z_{k}^{\Omega}]))\nonumber \\
 & = & \Gamma_{\Omega}(\Omega_{k}+b_{k}(g(\Omega(k),\alpha)+M_{k+1})),\label{eq:slow}
\end{eqnarray}

where $M_{k+1}=R_{\tau}z_{k}^{\Omega}-\mathbb{E}[R_{\tau}z_{k}^{\Omega}]$
is a zero mean martingale difference sequence with respect to the
$\sigma-$fields $\mathcal{F}_{t}=\sigma(\Omega_{n},\alpha_{n},N_{n},M_{n},n\leq t;t\geq0)$;
$g(\Omega(k),\alpha)=\mathbb{E}[R_{\tau}z_{k}^{\Omega}]$ and $\Gamma_{\Omega}:\mathbb{R}^{|N||m|}\rightarrow\mathbb{R}^{|N||m|}$
is the corresponding projection operator for $\Omega$ which ensures
that these iterates are projected to a compact region $W$ as in the
previous iterate update equation. We can thus represent the $\Omega$
update with the following ODE:

\[
\dot{\Omega}=\Gamma_{\Omega}(\mathbb{E}[R_{\tau}z_{k}^{\Omega}])
\]
Define the continuous time projection operators $\hat{\Gamma}_{\Omega}(v)=\lim_{\delta\rightarrow\infty}\frac{\Gamma_{\Omega}(\Omega+\delta v)-\Omega}{\delta}$
and $\hat{\Gamma}_{\alpha}(p)=\lim_{\delta\rightarrow\infty}\frac{\Gamma_{\alpha}(\alpha+\delta v)-\alpha}{\delta}$
that, given directions $v$ and $p$ to modify the paramaters $\Omega$
and $\alpha$ respectively ensures that the iterates are projected
into their compact sets $C$ and $W$ respectively. We can thus define
the ODEs using this continuous operator as:

\begin{align*}
\dot{\Omega} & =\hat{\Gamma}_{\Omega}(\mathbb{E}[R_{\tau}z_{k}^{\Omega}])\doteq\bar{g}(\Omega(k),\alpha)\\
\dot{\alpha} & =\hat{\Gamma}_{\alpha}(\mathbb{E}[R_{\tau}z_{k}^{\alpha}])\doteq\bar{f}(\alpha(k),\Omega)
\end{align*}
\textbf{Assumption (A1): }For each $\alpha\in\mathbb{R}^{d}$, the
ODE:

\[
\dot{\Omega}(t)=\bar{g}(\Omega(k),\alpha)
\]

has a globally asymptotically stable equilibrium $\bar{\lambda}(\alpha)$
such that $\bar{\lambda}:\mathbb{R}^{d}\rightarrow\mathbb{R}^{|N||m|}$
is Lipschitz.

\textbf{Assumption (A2): }The ODE:

\[
\dot{\alpha}=\bar{f}(\alpha(k),\bar{\lambda}(\alpha(k)))
\]

has a unique global asymptotically stable equilibrium $\alpha^{*}$.

In order to prove that these ODEs collectively converge, we need to
make the following assumptions.

\textbf{Assumption (A3): }The functions $f,g$ are Lipschitz continuous
functions

\textbf{Assumption (A4): }$\sup_{k}\Vert\Omega_{k}\Vert,\sup_{k}\Vert\alpha_{k}\Vert<\infty$

\textbf{Assumption (A5): }$\sum_{n}a(n)=\sum_{n}b(n)=\infty,$\textbf{
$\sum_{n}a(n)^{2}=\sum_{n}b(n)^{2}<\infty$}

\textbf{Assumption (A6): }For increasing $\sigma$-algebras, the martingale
sequences $\sum a_{k}N_{k},\sum b_{k}M_{k}<\infty$ a.s 

\textbf{Assumption (A7) }For all $\alpha,\Omega$, the objective function
$J(\mu_{\alpha,\Omega})$ has bounded second derivatives and the set
$Z$ of local optima of $J(\mu_{\alpha,\Omega})$ are countable.

Given the above assumptions, the parameter $\Omega_{k}\rightarrow\bar{\lambda}(\alpha^{*})$
and $\alpha_{k}\rightarrow\alpha^{*}$ as $k\rightarrow\infty$ a.s
by standard two-timescale stochastic approximation arguments \cite{Borkar1997}.
That is, the iterates converge to \{$\bar{\lambda}(\alpha^{*}),\alpha^{*}\vert\alpha^{*}\in Z\}$
.

\subsection{SARiCoS Video}
A video is attached with the supplementary material showing an agent (the striker) applying the learned risk-aware skills in a one-on-one scenario with a goalkeeper. The videos exhibit the Situational Awareness (SA) of the agent in both a losing scenario and a winning scenario.

\end{document}